\documentclass[letterpaper]{article} 
\usepackage{aaai24}  
\usepackage{times}  
\usepackage{helvet}  
\usepackage{courier}  
\usepackage[hyphens]{url}  
\usepackage{graphicx} 
\usepackage{natbib}  
\usepackage{caption} 
\usepackage{algorithm}
\usepackage{algorithmic}
\usepackage{amsmath}
\usepackage{xcolor}
\usepackage{multirow}
\usepackage{tabularx}
\usepackage{booktabs}
\usepackage{mathtools}
\usepackage{tikz}
\usepackage{amsfonts}  

\usepackage{pifont}
\newcommand{\cmark}{\ding{51}}%
\newcommand{\xmark}{\ding{55}}%

\usepackage[capitalize]{cleveref}
\crefname{section}{Sec.}{Secs.}
\Crefname{section}{Section}{Sections}
\Crefname{table}{Table}{Tables}
\crefname{table}{Tab.}{Tabs.}

\usepackage{newfloat}
\usepackage{listings}
\DeclareCaptionStyle{ruled}{labelfont=normalfont,labelsep=colon,strut=off} 
\floatstyle{ruled}
\newfloat{listing}{tb}{lst}{}
\floatname{listing}{Listing}
\title{
Trade-offs in Fine-tuned Diffusion Models Between Accuracy and Interpretability}
\author {
    Mischa Dombrowski\textsuperscript{\rm 1},
    Hadrien Reynaud\textsuperscript{\rm 2},
    Johanna P Müller\textsuperscript{\rm 1},
    Matthew Baugh\textsuperscript{\rm 2}, \\
    Bernhard Kainz\textsuperscript{\rm 1,2}
}
\affiliations {
    \textsuperscript{\rm 1}Friedrich-Alexander-Universität Erlangen-Nürnberg, DE\\
    \textsuperscript{\rm 2}Imperial College London, UK\\
    mischa.dombrowski@fau.de
}

\begin{document}

\maketitle

\begin{abstract}
Recent advancements in diffusion models have significantly impacted the trajectory of generative machine learning research, with many adopting the strategy of fine-tuning pre-trained models using domain-specific text-to-image datasets. Notably, this method has been readily employed for medical applications, such as X-ray image synthesis, leveraging the plethora of associated radiology reports. Yet, a prevailing concern is the lack of assurance on whether these models genuinely comprehend their generated content. With the evolution of text-conditional image generation, these models have grown potent enough to facilitate object localization scrutiny. Our research underscores this advancement in the critical realm of medical imaging, emphasizing the crucial role of interpretability. We further unravel a consequential trade-off between image fidelity -- as gauged by conventional metrics -- and model interpretability in generative diffusion models. Specifically, the adoption of learnable text encoders when fine-tuning results in diminished interpretability. Our in-depth exploration uncovers the underlying factors responsible for this divergence. Consequently, we present a set of design principles for the development of truly \emph{interpretable} generative models. Code is available at https://github.com/MischaD/chest-distillation.
 
%

\end{abstract}
%
%
%
\section{Introduction}
Automatic detection and localisation of diseases in medical images have great potential to be applied at scale because of the considerable amount of available data that connects images to radiology reports. 
Recently, multi-modal models have come into focus because of their ability to capture both, textual and visual information and combine them to improve  the performance of the model \cite{li2022groundedGLIP} and its interpretability \cite{huang2021gloria}.  
Additionally, \cite{boecking2022makingmscxr} showed how vision-language models can greatly benefit from prompt engineering because of the meaningful representations learned by large language models such as \cite{devlin2018bert,yan2022radbert,liu2020self}. 
Furthermore, these cross-modality abilities can also provide interpretable outputs, for example, by using contrastive learning approaches \cite{zhang2020contrastive,huang2021gloria,gupta2020contrastive} to perform phrase grounding, which associates certain tokens or words of the input prompt with regions in the image.
Due to recent advances in diffusion models, there has been an increased focus on generative approaches to solve common problems, such as categorical data imbalances and counterfactual image generation. 
Additionally, since their recent popularization by \citet{ho2020denoising}, diffusion models have propelled the performance of generative models \cite{dhariwal2021diffusion,saharia2021image,saharia2022photorealistic}. These advances have also been adopted in the medical domain, where most studies thus far have focused on improving generative capabilities, such as generating large corpora of MRI scans \cite{pinaya2022brain} or 4D data \cite{kim2022diffusion}. Current approaches have reached a level of performance where the models can generate videos of such high quality that clinicians cannot visually distinguish them from real videos \citep{reynaud2023featureconditioned}.
These successes also led to advances in discriminative tasks such as anomaly detection \cite{pinaya2022fast,wolleb2022diffusion}. 
One common approach is to use pre-trained diffusion models~\cite{Rombach_2022_CVPR} and fine-tune them to generate, for example, Chest X-ray (CXR) images. 
The resulting image quality is better compared to training diffusion models from scratch, as demonstrated in \cite{chambon2022roentgen}. 
These approaches have in common that they do not interpret the results given by the diffusion models, which have recently demonstrated, particularly, interpretable latent spaces \cite{dombrowski2022zero,tang2022daam,hertz2022prompt}. 
Interpreting the latent space of diffusion models is of great importance because generative models have to properly represent what the tokens mean in order to generate correctly matching images from them. 
Even if we validate the generated samples on pre-trained classifiers~\cite{chambon2022roentgen}, the results can be deceiving. 
Many models have not been evaluated on robustness towards unrealistic artefacts introduced by generative models. 
Furthermore, if classifiers are already able to clearly label images as belonging to a class, it is unclear whether this sample is useful for applications such as data augmentation.

In this paper, we show that the state-of-the-art methods for fine-tuning diffusion  models on radiology reports produce models which are no longer interpretable. 
We do this by analyzing the influence of jointly learning language embedding and image generation. 
Our experiments show that diffusion models trained for high image quality have no semantic understanding of their conditional input, but instead, learn to produce images based on confounders.  
Our hypothesis is that jointly learning the text and image embedding is too difficult and that adapting language models to medical tasks should be approached with more care, including a thorough analysis of the interpretability capabilities of the diffusion models.
To show this, we evaluate the phrase grounding potential of the diffusion model on the state-of-the-art approach to fine-tune diffusion models. 
Then we empirically analyze the effect of keeping the textual encoder frozen. 
This results in statistically inferior generative performance, but retains the interpretability of the model to such an extent that we beat discriminative phrase grounding approaches on two out of eight diseases in a selected clinical downstream scenario.
Furthermore, we statistically evaluate the significance of our findings. 
By demonstrating the effect of losing accuracy to gain interpretability, this paper is the first work that unveils this important trade-off within the domain of generative medical imaging models.

\section{Method}
\label{sec:Method}
\begin{figure*}[t]
    \centering
    \includegraphics[width=\linewidth]{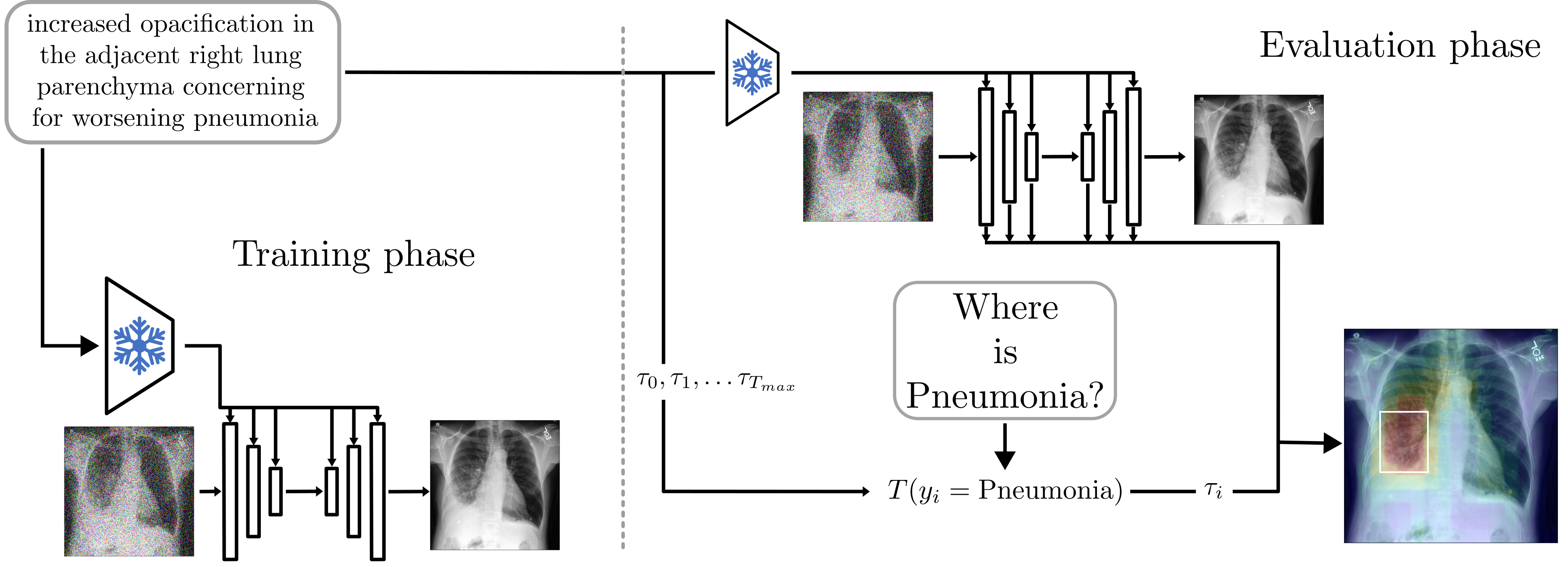}
    \caption{Visual summary of our findings. We investigate different ways of finetuning multi-modal diffusion models. We show that by keeping the language encoder frozen, visualized by the blue snowflake, we can train diffusion models that are interpretable by evaluating the localization abilities of the attention map.}
    \label{fig:visual_abstract}
\end{figure*}

\noindent\textbf{Pretrained foundation model} 
We begin with Stable Diffusion v2 (SDv2) \cite{Rombach_2022_CVPR} as a pre-trained foundation model. 
Using pre-trained models as a starting point is a common approach to improve training time and, furthermore, \cite{chambon2022roentgen} have shown that it produces better results in terms of image quality compared to starting from a randomly initialized model. 
SDv2 is a latent diffusion model, which means that the diffusion is learned on a reduced image size $Z = 64$. Following \cite{chambon2022roentgen}, we keep the model used to compute the latent dimension fixed. 
Consequently, we can precompute the latent representation of our input images to speed up the learning time and to fit the entire dataset into system memory.  
Since SDv2 is a text-to-image diffusion model, it was trained to learn the function $f(y)$ that generates images conditioned on textual input $y$. At training time, this prompt could be ``a photo of a swiss mountain dog''. 
In order to inject the conditional input into the diffusion model, the text is first tokenized using a tokenizer $T(y)$ and then embedded using a clip-like model \cite{radford2021learning}, in this case, a ``ViT-H-14'' \cite{dosovitskiy2020image}. The tokenizer maps words to integers as input for the textual encoder and has a fixed vocabulary size determined at training time. Short input prompts are padded to a fixed length of $T_{max}$, which is typically set to 77 at training and inference time.
Since the language embedder was trained on normal images and captions, many words from the medical domain do not appear in this vocabulary. 
The tokenizer solves this issue by splitting unknown words into multiple tokens. 
``Consolidation'' for example exists, whereas the word ``Cardiomegaly'' is split into four tokens. 
To formalize this, we define $\tau_{i} \coloneqq {\forall \tau : \tau \in T(y_i)}$ as the set of tokens corresponding to input word $i$.
In the next step, all tokens are encoded using a pre-trained language encoder. 
It is common to use a learnable language encoder to simultaneously train the language encoder and the diffusion model, as this results in higher image quality \cite{chambon2022roentgen}. 

\noindent\textbf{Classifier-free guidance}
For image synthesis, we use classifier-free guidance, which has been shown to produce better results in terms of image fidelity in the medical domain \cite{chambon2022roentgen}. Intuitively, this means that the reverse diffusion step is performed twice, once for unconditional denoising and once for conditional denoising. Then the difference is computed and amplified to push the denoising step further into the direction of the conditioning and improve the quality of the results. To improve the performance of conditional denoising, we do unconditional guidance training, a common technique that drops the conditional input for healthy samples. We executed this in 30\% of the ``No Finding'' cases.  

\noindent\textbf{Interpretability}
Following \citet{zhang2018visual} we claim that interpreting mid-layer representations of deep neural networks helps to interpret the predictions of the model. In the case of text-conditioned diffusion models this means that we analyze the interpretability by looking at their attention layers following the approach proposed by \citet{dombrowski2022zero}. 
Internally, every input token is converted to a learned latent embedding. This embedding is inserted into the model at every level of the denoising U-Net. The denoising U-Net itself consists of multiple channels and pixels with a reduced width and height. 
Attention is computed at every layer as the dot-product of the latent representation of the image with 
the embedding of the token: 
\begin{equation}
    \text{Attention}(Q, K, V) = \text{softmax}(\frac{QK^T}{\sqrt{d}}) \cdot V
\end{equation}
where K and V are the learned projection representations of the input token, Q is the learned projection of the input image, and d is a normalization constant.  
This means that the layers produce output conditioned on input to localize the relevant pixel-wise features. 
In the next step we can  save the attention maps of multiple reverse diffusion steps of all cross-attention layers of the U-Net architecture and compute the mean over it. 
Attention maps in downsampled layers are resized to the size $Z \times Z$.
The resulting attention map has the shape $\mathbb{R}^{T_{max} \times Z \times Z}$. 
Due to the softmax operation, attention is defined as a probability and the values over T$_{max}$ sum up to one. 
To get a class prediction, all we have to do, is to compute the location of all tokens $\tau_i$ corresponding to this class and compute the average. 
For in-domain data, this is straightforward as all the words have a single token representation due to them being part of the training dataset of the model. 
For unknown words, the tokenizer splits the words into several known tokens. 
Therefore, we aggregate the attention over all tokens corresponding to this word. 
The pipeline to analyze this type of interpretability is summarized in \cref{fig:visual_abstract}. 

\section{Datasets}
\noindent\textbf{In-domain data}
To fine-tune a diffusion model on in-domain data we take MS-COCO \cite{lin2014microsoft}, a giant dataset comprised of natural images in their usual environment, and fine-tune on the captions of these images. 
We split the official training set into 98169 train and 10702 validation images. 
We use the official validation set as our testset as we do not have ground truth segmentation labels available for the test set. 
If multiple captions are available for a single image, we use all captions for training but randomly sample one of them for the training steps. 
During validation we first filter the dataset such that only images remain where the object class is part of the caption to make the task of token-based localization easier. For example, for the class of birds ``two seagulls standing on a handrail'' would be removed but ``two birds standing on a hand rail'' would be left in. 
Since these captions are short descriptions of the content of the image this step introduces a bias towards only evaluating our model on images where the object is prominent.

\noindent\textbf{Out-of-domain data} To fine-tune the models on data from a different domain, we use posterior-anterior (PA) and anterior-posterior (AP) views of MIMIC-CXR \cite{johnson2019mimic}, a large corpus of paired text and image CXR data. For evaluating localization results we use the phrases and bounding boxes of MS-CXR, a subset of MIMIC-CXR with revised bounding boxes and impressions curated by board-certified radiologists based on the original reports \cite{boecking2022makingmscxr} and therefore serve as a higher quality textual description of the image content.
Evaluation of multi-label images is done by splitting the image into two instances, one for each disease class. 
To evaluate segmentation results, we do the following preprocessing step: 
If the image has multiple bounding boxes describing the same disease, we keep them together and provide the results for the union of multiple bounding boxes. For phrase grounding benchmarks, these are kept separate if the phrases for the bounding boxes are different so that we can compare our method to other methods. 
To evaluate the statistical significance of our findings we report the average results over five runs.
Following \cite{chambon2022roentgen}, we only considered samples with impressions that have over 7 characters for training. 
MS-CXR and the p19 subset of MIMIC are left out as a test set which results in a total of 162651 samples for the training set. 
The p19 subset is used to test the performance of our proposed image generation method. We sample 5000 images from this subset, reducing the number of ``No Finding'' samples to avoid reporting our generative metrics on a subset consisting of predominantly healthy images. 

\subsection{Evaluation metrics}
To evaluate feature localization accuracy we use AUC-ROC, Top-1 accuracy and contrast to noise ratio (CNR) \cite{boecking2022makingmscxr} as they allow quantifying the performance without tuning thresholds. 
Top-1 in this context means the accuracy of the highest predicted pixel being within the boundary of the ground truth annotation. 
Intuitively this score is supposed to caption the attention of the human observer who looks at the output and focuses on the highest prediction value first. 
In practice, this value is weak against outliers, \emph{e.g.} if annotations in the corner of the image are detected as outliers.
To assess the generative performance we report the Fréchet inception distance (FID) of 5000 images using the standard InceptionV3 and additionally a domain-specific DenseNet121 from \cite{Cohen2022torchxrayvision} (FID$_{XRV}$) for the CXR data. The images used for comparison are first resampled from the p19 test set to limit the number of ``No Finding'' input conditions (For details see Supplements). The sampling settings for the diffusion model follow \cite{chambon2022roentgen}.
To assess the diversity of our samples, we use MS-SSIM \cite{wang2003multiscalessim} on pairs of 4 images created with the same prompt, averaged over a set of 100 prompts.

\subsection{Baseline Method}
As the baseline method, we use Stable Diffusion v2, a text-to-image model trained for image generation of natural images of size $512 \times 512$ \cite{Rombach_2022_CVPR}. 
We perform all our fine-tuning experiments closely following the recommendations from \cite{chambon2022roentgen} (Finetune$_{L}$), by choosing a learning rate of $5 \times 10^{-5}$, a batch size of 256 and training for 60000 steps. When sampling, we use a classifier-free guidance scale of 4 over 75 sampling steps using the PNDM sampler \cite{liu2022pseudo}. When keeping the language encoder frozen (Finetune$_{F}$), we observe that the models perform better after only 30000 training steps, due to the simplified training objective and fewer parameters. We put ablations on this in the supplementary material.
Our approach differs from \cite{chambon2022roentgen} because we do not exclude AP views from our evaluation. The reason behind this is that the MS-CXR test set contains both AP and PA views. Furthermore, we refrain from limiting the number of healthy samples in our training set. 
All results are calculated after applying resizing and center-cropping to get to an image size of $512 \times 512$.
We split our fine-tuning over 16 80GB A100 GPUs which takes roughly 240 GPU hours with the frozen, and 580 with the learnable language encoder.

\section{Results}


\begin{table*}
\centering

\begin{tabularx}{\linewidth}{Xcccp{0.0cm}cccp{0.0cm}ccc} 
\toprule
\multirow{2}{*}{Category} & \multicolumn{3}{c}{SDv2} &&\multicolumn{3}{c}{ Finetune$_{L}$ } && \multicolumn{3}{c}{ Finetune$_{F}$ (Ours) } \\
\cmidrule(lr){2-4}\cmidrule(lr){6-8}\cmidrule(lr){10-12}
& AUC $\uparrow$  & Top-1 $\uparrow$ & CNR $\uparrow$ && AUC $\uparrow$  & Top-1 $\uparrow$ & CNR $\uparrow$ & &AUC $\uparrow$  & Top-1$\uparrow$  & CNR $\uparrow$ \\
\cmidrule(lr){2-2}\cmidrule(lr){3-3}\cmidrule(lr){4-4}\cmidrule(lr){6-6}  \cmidrule(lr){7-7}\cmidrule(lr){8-8} \cmidrule(lr){10-10} \cmidrule(lr){11-11}\cmidrule(lr){12-12}
Atelectasis       & 58.5   & 0.0  & 0.23  && 45$\pm$4.9  & 2$\pm$1.6   &  -0.15$\pm$.13  &&  \textbf{86$\pm$0.9}  &  \textbf{80$\pm$3.3} & \textbf{1.25 $\pm$.06}  \\
Cardiomegaly      & 53.1   & 8.4  & 0.14  && 45$\pm$1.6  & 2$\pm$0.8   &  -0.16$\pm$.05  &&  \textbf{73$\pm$2.3}  &  \textbf{40$\pm$6.8} & \textbf{0.67 $\pm$.07}  \\
Consolidation     & 59.3   & 2.6  & 0.21  && 53$\pm$4.8  & 14$\pm$4.3  &  0.05 $\pm$.11  &&  \textbf{86$\pm$1.6}  &  \textbf{48$\pm$7.5} & \textbf{1.22 $\pm$.10}  \\
Edema             & 80.2   & 15.2 & 0.37  && 54$\pm$1.6  & 29$\pm$8.2  &  0.09 $\pm$.02  &&  \textbf{89$\pm$0.1}  &  \textbf{70$\pm$3.8} & \textbf{1.36 $\pm$.01}   \\
Lung Op.          & 64.7   & 1.2  & 0.87  && 53$\pm$3.8  & 7$\pm$4.6   &  0.05 $\pm$.09  &&  \textbf{83$\pm$0.2}  &  \textbf{38$\pm$1.9} & \textbf{1.06 $\pm$.01}  \\
Pl. Effusion      & 48.8   & 1.0  & 0.34  && 40$\pm$4.7  & 3$\pm$1.6   & -0.27$\pm$.12   &&  \textbf{83$\pm$1.1}  &  \textbf{66$\pm$3.8} & \textbf{1.05 $\pm$.06}  \\
Pneumonia         & 61.3   & 0.5  & 0.64  && 49$\pm$2.4  & 3$\pm$1.9   & -0.05$\pm$.05   &&  \textbf{85$\pm$0.9}  &  \textbf{59$\pm$4.5} & \textbf{1.16 $\pm$.04}  \\
Pneumothorax      & 71.0   & 10.2 & 0.01  && \textbf{81$\pm$6.9}  & \textbf{27$\pm$26.5} & \textbf{0.97 $\pm$.33}   &&  74$\pm$2.7  &  20$\pm$3.4 & 0.68 $\pm$.08  \\

\midrule                                                                                              
Average           & 60.6   & 5.7  & 0.321   && 54$\pm$3.7  & 10$\pm$7.4  & 0.133$\pm$.12  &&  \textbf{79$\pm$0.5}  & \textbf{45$\pm$1.5}  & \textbf{0.920$\pm$0.03} \\
\bottomrule
\end{tabularx}
\caption{AUC-ROC and Top-1 accuracy for the phrase grounding benchmark of different chest diseases on MS-CXR. We compare our approach compared to SDv2 \citep{Rombach_2022_CVPR} and Finetune$_{L}$ \citep{chambon2022roentgen}. Mean and standard deviation over three different trainings are given. The results show that training with a frozen language encoder significantly improves the localization performance throughout all metrics for seven out of eight disease clasess.}

\label{tab:segmentationconditional}
\end{table*}
\subsection{In-domain Fine-tuning}
First, we evaluate our method with respect to segmentation results on MS-COCO data. We argue that this data is much closer to the training dataset of SDv2 as this model was trained to be a general-purpose text-to-image model. 
Therefore, we first analyze whether the finetuning process itself destroys the ability of the generative model to be interpretable, especially after finetuning them on smaller compute clusters.
The original model \cite{Rombach_2022_CVPR} was trained on a batch size of 2048, which enables the model to learn a joint representation of the text and the image and therefore enables interpretation of the latent space \cite{dombrowski2022zero}. In \cref{tab:mscoco_finetuning_length} we show, that finetuning on in-domain data remains possible even on a single Nvidia A100 GPU and a batch size of 16. 

\noindent\textbf{Trade-off of learnable vs frozen encoder}
Interestingly, we can see that the FID improves over time for both the learnable and the frozen encoder. However, in the case of the learnable encoder, the task of jointly learning language embedding and conditional image generation was too hard and therefore, the FID got worse first. We can also see this in terms of interpretability measured in AUC-ROC. The initial value is way worse than the pretrained baseline and the frozen model. However, eventually the interpretability and synthetic quality get better. 
The Frozen finetuning works better, with an improved FID score compared to the baseline even for a low number of steps. Further finetuning the model leads to the best FID score and a small improvement in terms of AUC-ROC. 
Interestingly, in terms of segmentation, the pretrained SDv2 model is the best out of all of them. Partially, this can be explained by the training dataset.
SDv2 was trained on a subset of the LAION-5B dataset \cite{schuhmann2022laion}, which is a huge agglomeration of natural images with text descriptions scraped from the internet. It is likely that a huge chunk of MS-COCO is already part of the training data.
Additionally, the huge variability in the training dataset of SDv2 helps the model to understand concepts better and the result is marginally higher interpretability, even if we compare it to a model finetuned to a more specific dataset.

\begin{table}
    \centering
    \begin{tabular}{l|X|c|c|c|c}
    Method & Steps & Frozen & FID $\downarrow$ & AUC-ROC $\uparrow$ \\
    \toprule
    SDv2                      & 0 & -  & 26.1 & \textbf{91.8}\\
    \midrule
    \multirow{4}{*}{Finetune} &  10000  & \xmark & 29.5     &  79.0 \\
                              &  150000 & \xmark & 25.0    & 90.1  \\
                              &  10000  & \cmark & 25.8    &  91.3 \\
                              &  150000 & \cmark & \textbf{19.4} &91.5 \\
    \bottomrule
    \end{tabular}
    \caption{MS-COCO finetuning experiment: We compare the generative and object segmentation performance of the two suggested finetuning models. Keeping the language encoder frozen leads to better interpretability and faster convergence.}
    \label{tab:mscoco_finetuning_length}
\end{table}

\subsection{Localization Results}
Next, we want to assess the interpretability of these models when training them on data of new domains. We do this by looking at localization metrics computed on the impressions from MS-CXR \cite{boecking2022makingmscxr}.
Choosing tokens $\tau_{i}$ to attend to is not straightforward for textual conditioning on medical data. 
Hence, in order to compare with phrase-grounding approaches, if the name of the disease, or some altered version of it appears in the impression, we use the attention maps of this token as the prediction.  
If this does not occur, we compute the attention for the tokens of all the words, excluding tokens indicating the start of a string, the end of a string, and padding.
Manually choosing tokens in this case, \emph{e.g.}, the token for ``heart'' to localize Cardiomegaly could potentially boost the performance even further, but we avoid this for the sake of generalizability. 
Since we observed limited localization results for the learnable approach, we do not compute the absolute value of CNR in order to not wrongly overestimate the performance. 

Quantitative results are shown in \cref{tab:segmentationconditional}. 
SDv2 generally has a bad localization performance, although some classes had better scores than expected. One possible explanation is that some diseases have correlations with certain regions, like the regions with medical equipment, lungs, or text labels, that SDv2 often puts higher attention to. 

Finetune$_L$ is not able to perform localization consistently.
Its localization performance only reaches 54\% AUC-ROC which is even worse than the SDv2 baseline and 10\% Top-1 accuracy, which is slightly better. 
Overall, this indicates that this model has not acquired any understanding of the disease-defining features. 
Interestingly, the localization for ``Edema'', ``Consolidation'' and ``Pneumothorax'' has a decent Top-1 Accuracy and the CNR results for ``Pneumothorax'' shows the best AUC-ROC out of all methods. However, the large standard deviation indicates that the understanding of this disease is not robust and might be related to random artefacts.

\begin{figure}[t]
    \centering
    \includegraphics[width=.95\linewidth]{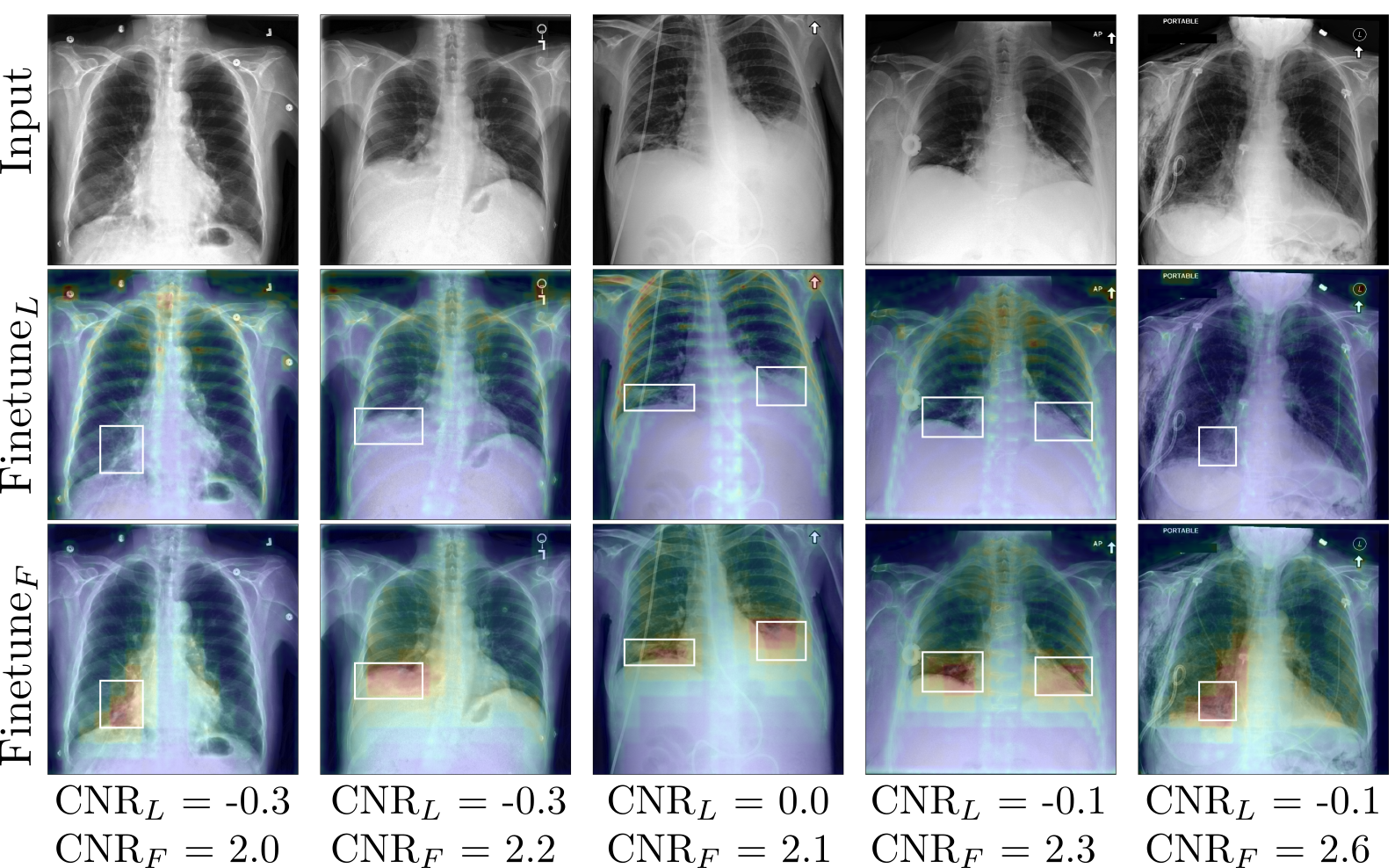}
    \caption{Phrase grounding examples from MS-CXR. White labels show which bounding boxes are ground truth. Pixels with higher importance are shown in red.}
    \label{fig:segmentation_comparison}
\end{figure}

\noindent\textbf{Trade-off of learnable vs frozen encoder}
Keeping the language encoder frozen during fine-tuning, on the other hand, shows excellent results in terms of localization. 
It achieves 79\% AUC-ROC, 45\% Top-1 accuracy, and 0.920 CNR across all diseases. 
The attention maps are therefore satisfactory indicators for the location of the disease, and we infer that the model has learned to localize those conditions.
\cref{fig:segmentation_comparison} confirms our qualitative observations by showing a considerable gap in the interpretability of the two different diffusion models. Finetune$_L$ mistakenly relates $\tau_{i}$ to image features, such as rips, spine, or bones, that are unrelated to the disease class. Our proposed approach, on the other hand, shows a good localization performance, including multi-instance input samples. 

\begin{table*}[ht]
\begin{tabularx}{\textwidth}{lcXlccccccccc}
\toprule
 \multirow{5}{*}{\rotatebox{90}{Weakly}} & \multirow{5}{*}{\rotatebox{90}{Supervised}} & Method  & \multicolumn{1}{l}{Atel.} & \multicolumn{1}{l}{Card.} & \multicolumn{1}{l}{Cons.} & \multicolumn{1}{l}{L-Op.} & \multicolumn{1}{l}{Edem.} & \multicolumn{1}{l}{Pnem.} & \multicolumn{1}{l}{Pnth.} & \multicolumn{1}{l}{P-Ef.} & \multicolumn{1}{l}{Avg.} \\ 
\cmidrule(lr){4-4} \cmidrule(lr){5-5}\cmidrule(lr){6-6}\cmidrule(lr){7-7}\cmidrule(lr){8-8}\cmidrule(lr){9-9}\cmidrule(lr){10-10}\cmidrule(lr){11-11}\cmidrule(lr){12-12} 
& & ConVIRT \cite{zhang2020contrastive}   & 0.86   & 0.64   & 1.25   & 0.78   & 0.68   & 1.03   & 0.28   & 1.02   & 0.818 \\
& & GLoRIA  \cite{huang2021gloria}  & 0.98   & 0.53   & 1.38   & 1.05   & 0.66   & 1.18   & 0.47   & 1.20   & 0.930 \\
& & BioViL-L \cite{boecking2022makingmscxr} & 1.17   & \textbf{0.95}   & \textbf{1.45}   & \textbf{1.19}   & 0.96   & \textbf{1.19}   & 0.74   & \textbf{1.50}  & \textbf{1.142} \\ \midrule
 \multirow{3}{*}{\rotatebox{90}{Zero-}} & \multirow{3}{*}{\rotatebox{90}{shot}} & SDv2  \cite{Rombach_2022_CVPR} & 0.23   & 0.14   & 0.21   & 0.37   & 0.87   & 0.34   & 0.64   & 0.01   & 0.321 \\ 
& & Finetune$_{L}$  \cite{chambon2022roentgen}& -0.15  & -0.16  & 0.05  & 0.05   & 0.09  & -0.05  & \textbf{0.97}   & -0.27  & 0.133 \\ 
& & Finetune$_{F}$ (Ours)& \textbf{1.25}   & 0.67   & 1.22   & 1.06   & \textbf{1.36}   & \textbf{1.16}   & 0.68   & 1.05   & 0.920 \\ 
\bottomrule
\end{tabularx}
\caption{Phrase grounding results for contrast-to-noise ratio (CNR) on MS-CXR of our approach compared to state-of-the-art weakly supervised phrase grounding results. Values for ConVIRT, GLoRIA, and BioViL-L are taken from \cite{boecking2022makingmscxr}. Despite being generative in nature, interpreting the attention layers of the diffusion model and using them for phrase grounding reaches a similar performance as weakly supervised approaches.}
\label{tab:cnr}
\end{table*}

\noindent\textbf{Phrase grounding benchmarks}
To substantiate the performance of our method, we can compare it with phrase grounding benchmarks. These methods were trained using contrastive methods and were specifically designed with the task of localization in mind.
Our method, on the other hand, provides these localization maps without additional effort, and they can therefore be used for localization in a zero-shot manner. 
As it can be seen in \cref{tab:cnr}, 
our best method outperforms the discriminative approaches in terms of localization in two out of eight disease classes, which is remarkable, given that our model is \emph{only} generative by nature.
The learnable method, once again, is the worst method out of all generative approaches, with slightly worse results than our SDv2 baseline.

\begin{table*}
\centering

\begin{tabularx}{\linewidth}{p{5.0cm}XcXccXcc} 

\toprule
\multirow{2}{*}{Method}   & \multicolumn{2}{c}{FID $\downarrow$}  & &\multicolumn{2}{c}{FID$_{XRV}$ $\downarrow$ }  &&\multicolumn{2}{c}{MS-SSIM $\downarrow$}  \\
\cmidrule(lr){2-3}\cmidrule(lr){5-6}\cmidrule(lr){8-9}
                & Mimic & MS-CXR && Mimic & MS-CXR &&Mimic & MS-CXR \\
\cmidrule(lr){2-2}\cmidrule(lr){3-3}\cmidrule(lr){5-5}\cmidrule(lr){6-6}\cmidrule(lr){8-8}\cmidrule(lr){9-9}
SDv2 \cite{Rombach_2022_CVPR}            & 237.6& 236.2 && 104.9 & 109.4  && 12.9 & 11.1 \\
Finetune$_{L}$ \cite{chambon2022roentgen} & \textbf{61.9} & \textbf{60.5}  & &\textbf{7.7}   & \textbf{7.3}  & & 10.3 & 11.4 \\
Finetune$_{F}$ (Ours) & 75.5 & 75.7  && 10.1  & 10.0 & &\textbf{10.1} & \textbf{8.2}  \\
\bottomrule
\end{tabularx}
\caption{Quantitative comparison of fidelity in terms of general FID and domain-specific FID$_{XRV}$ score and diversity of image generation in terms of MS-SSIM (in \%). We compare the results obtained from using impressions from Mimic or from MS-CXR as our prompts for generation.}
\label{tab:unconditionalgeneration}
\end{table*}

\subsection{Generative Results}
\label{sec:generative_quality}
Next, we evaluate the generative quality of the different models. 
Table \ref{tab:unconditionalgeneration} shows the quantitative results of the fine-tuning. 
The fidelity scores of the models follow the observations reported in \cite{chambon2022roentgen} that the results are drastically improved by making the textual encoder learnable during fine-tuning, despite the evidence that the model has a worse understanding of what individual tokens mean.
To investigate the correctness of the conditional generation, we evaluate the prediction AUC-ROC of a pre-trained classifier, similar to the approach chosen by \cite{chambon2022roentgen}. 
Even though no ground-truth labels are available for the generated images, it indicates whether the generated artefacts resemble the ones used by trained classifiers.
For that, we use the label predictions of the previously used DenseNet121 on our generated images and compare them with the predictions of the same label for the generated healthy images. 
As a baseline, we take the classification accuracy on 5000 real images from the p19 subset.
Table \ref{tab:condimagegen} shows the results. We can conclude that the class conditional generation follows the same trend as the unconditional generation. The best model is the one that jointly trains language embedding and image generation. 
However, the difference between both models is very small compared to the FID scores. In fact, Finetune$_F$ even outperforms the other method in six out of eight classes. 
The only exception  are ``Pneumothorax'', which was also the only disease class that had good AUC-ROC performance, and ``Edema'', which was the only class with good Top-1 accuracy, as shown \cref{tab:segmentationconditional}. 

\begin{table*}

\begin{tabularx}{\linewidth}{p{4.8cm}Xccccccccc}
\toprule
Method   & Source &   \multicolumn{1}{c}{Atel.} & \multicolumn{1}{c}{Card.} & \multicolumn{1}{c}{Cons.} & \multicolumn{1}{c}{L-Op.} & \multicolumn{1}{c}{Edem.} & \multicolumn{1}{c}{Pnem.} & \multicolumn{1}{c}{Pnth.} & \multicolumn{1}{c}{Effu.} & \multicolumn{1}{c}{Avg.} \\ 
\cmidrule(lr){3-3}\cmidrule(lr){4-4} \cmidrule(lr){5-5}\cmidrule(lr){6-6}\cmidrule(lr){7-7}\cmidrule(lr){8-8}\cmidrule(lr){9-9}\cmidrule(lr){10-10}\cmidrule(lr){11-11} 
p19 Test & Real  & 83.0 & 85.5 & 87.7 & 80.2 & 91.4 & 75.6 & 84.4 & 88.7 & 84.6 \\ \midrule
Finetune$_L$ \cite{chambon2022roentgen} & \multirow{2}{*}{Synthetic} & 75.8 & 78.7 & \textbf{77.8} & 75.1 & \textbf{85.9} & 66.2 & \textbf{79.6} & 79.6 & \textbf{77.3} \\ 
Finetune$_{F}$ & & \textbf{76.8} & \textbf{79.0} & \textbf{77.9} & \textbf{77.5} & 84.1 & \textbf{66.7} &   70.1 & \textbf{84.5} & 77.1 \\ 
\bottomrule
\end{tabularx}
\caption{AUC-ROC of a pre-trained classifier \cite{Cohen2022torchxrayvision} on real samples (p19) compared to the accuracy on synthetic images. The metrics are computed as the average over the scores from all diseases. Results for Pleural Effusion are reported using the prediction for Effusion since the classifier does not distinguish between different types of effusions.}
\label{tab:condimagegen}
\end{table*}


\noindent\textbf{Lack of interpretability}
To investigate where this lack of interpretability comes from
we look for every word  that occurs more than 25 times in the test set, how good the localization is and report the three best words and values for all three models in \cref{tab:MSCXRtokens}. 
Interestingly, we can see that Finetune$_L$ has good values for some of the words in the dataset but these words are never consistent with the diseases. The only exception is the end of string token which has a good localization performance throughout all diseases. Due to the language encoding, which is causal, tokens in the future are conditioned on tokens before that. 
Therefore, the end of string token has learned to relate the text to the location in which the important information is located.   
Overall the end-of-string (EOS) token of Finetune$_L$ has a CNR of 0.68 which is still a bit worse than EOS for Finetune$_F$ with 0.72 and far from the performance of Finetune$_L$ conditioned on the disease describing words.  
For phrase grounding this method of localization is not feasible as it fails if parts of the phrase describe different regions of the image.

\begin{table*}
\centering
{\begin{tabularx}{\linewidth}{Xccccccccccc} 
\toprule
 \multirow{2}{*}{Category} & \multicolumn{3}{c}{SDv2 \cite{Rombach_2022_CVPR}}      &&\multicolumn{3}{c}{ Finetune$_{L}$ \cite{chambon2022roentgen}} && \multicolumn{3}{c}{ Finetune$_{F}$ }\\
 \cmidrule(lr){2-4}\cmidrule(lr){6-8}\cmidrule(lr){10-12}
  & Word  & Occ. & CNR &&  Word  & Occ. & CNR &&  Word  & Occ. & CNR \\


 \midrule
 \multirow{3}{*}{Atelactasis}& EOS  & 61 & 0.37 &&  EOS & 61& 0.80 && atelectasis &  50& 1.11 \\ 
 & atelectasis  & 50 & 0.25 &&  SOS & 61& -0.17 && EOS &  61& 0.85 \\ 
 & SOS  & 61 & -0.17 &&  atelectasis & 50& -0.23 && SOS &  61& -0.85 \\ 
 \midrule 
\multirow{3}{*}{Cardiomegaly}& EOS  & 333 & 0.24 &&  EOS & 333& 0.37 && enlarged &  222& 1.16 \\ 
 & cardiac  & 220 & 0.21 &&  SOS & 333& 0.19 && silhouette &  219& 0.92 \\ 
 & silhouette  & 219 & 0.16 &&  moderate & 30& -0.09 && moderate &  30& 0.84 \\ 
 \midrule 
\multirow{3}{*}{Consolidation}& lung  & 56 & 0.82 &&  EOS & 117& 0.99 && lung &  56& 1.41 \\ 
 & lobe  & 32 & 0.61 &&  patchy & 60& 0.11 && consolidation &  93& 1.34 \\ 
 & patchy  & 60 & 0.58 &&  left & 44& 0.06 && left &  44& 1.26 \\ 
 \midrule 
\multirow{3}{*}{Edema}& pulmonary  & 39 & 1.20 &&  EOS & 46& 0.90 && pulmonary &  39& 1.25 \\ 
 & EOS  & 46 & 0.73 &&  pulmonary & 39& 0.18 && edema &  42& 1.20 \\ 
 & edema  & 42 & 0.53 &&  edema & 42& 0.13 && EOS &  46& 0.96 \\ 
 \midrule 
   \multirow{3}{*}{Lung opacity}& lung  & 35 & 0.75 &&  EOS & 82& 0.97 && lung &  35& 1.33 \\ 
 & ground-glass  & 32 & 0.60 &&  the & 43& 0.10 && EOS &  82& 1.06 \\ 
 & patchy  & 33 & 0.55 &&  right & 32& 0.09 && opacity &  28& 1.00 \\ 
 \midrule 
\multirow{3}{*}{Effusion}& SOS  & 96 & 0.11 &&  EOS & 96& 0.52 && left &  31& 0.84 \\ 
 & EOS  & 96 & 0.03 &&  SOS & 96& 0.03 && pleural &  88& 0.74 \\ 
 & left  & 31 & -0.12 &&  Small & 39& -0.20 && effusion &  51& 0.71 \\ 
  \midrule
\multirow{3}{*}{Pneumonia}& lung  & 46 & 0.72 &&  EOS & 182& 0.89 && lung &  46& 1.33 \\ 
 & lobe  & 49 & 0.68 &&  right & 58& -0.01 && consolidation &  31& 1.26 \\ 
 & EOS  & 182 & 0.45 &&  opacities & 30& -0.01 && lobe &  49& 1.26 \\ 
 \midrule
 \multirow{3}{*}{Pneumothorax}& small  & 70 & 0.73 &&  small & 70& 0.95 && small &  70& 0.70 \\ 
 & pneumothorax  & 226 & 0.58 &&  left & 85& 0.70 && apical &  119& 0.66 \\ 
 & apical  & 119 & 0.54 &&  apical & 119& 0.69 && pneumothorax &  226& 0.46 \\ 

\bottomrule
\end{tabularx}
}
\caption{Number of occurrences and CNR for the three words with the highest localization accuracy of each model on MS-CXR. SOS and EOS are start and end of string respectively. The EOS string token shows consistently high localization performance for both fine-tuned models. However, the frozen model is the only model that learns the connection between the token describing the disease and its localization.
}
\label{tab:MSCXRtokens}
\end{table*}

\subsection{Influence of Complexity of Text Input}
Finally, we want to understand the influence of the text data on the interpretability.  Given the results of \cref{tab:segmentationconditional} we hypothesize that the learnable models struggle with the large variability and the complexity of the impressions of the radiology report in the training datasets. Therefore, we experiment with the variability of the textual input phrases. Our goal is to maximize the interpretability of the model by directly injecting the class condition into the text prompts assuming that we can boost phrase grounding performance. 
The highest variability is obtained by taking the impressions directly (compare \cref{tab:segmentationconditional}. The lowest variability is obtained by taking the class, converting it to a string, and taking this as input. Additionally, to get a text dataset with medium variability, we take ChatGPT version 3.5 and ask it to generate 100 radiography impressions of CXR images with a certain disease (We show a few examples in the supplementary material). Then we randomly sample from this set during training. To enable a fair comparison we only report the results on the subset of the MS-CXR test dataset that has the label of the image in its impression. 
The results are given in \cref{tab:labelandchatgpttrainign}. 

\noindent\textbf{Trade-off of learnable vs frozen encoder}
The learnable models once again completely fail in terms of interpretability. We achieve the highest phrase grounding performance when training the model on the impressions directly. Training on labels surprisingly does not improve the localization results. We assume this is partial because of the domain gap in the text input when we try our localization method on impressions that not only consist of the class label itself. The synthetic impressions achieved the lowest result validating our observation that a high variability of the impressions is key for the model to be interpretable.

\begin{table}
    \centering
    \begin{tabular}{c|c|cc}
    \toprule
    \multirow{2}{*}{Textual input} & \multirow{2}{*}{Variability} & \multicolumn{2}{c}{CNR}\\
                                   &                              & Learnable & Frozen \\
    \midrule
    Impressions      & high   &0.15& \textbf{0.95} \\
    ChatGPT          & middle &0.22& 0.42          \\
    Class labels     & low    &0.29& 0.80          \\
    \bottomrule
    \end{tabular}
    \caption{CNR results for different models in relation to the variability of the textual input prompts. Maximizing the variability of the prompts also maximizes the interpretability.}
    \label{tab:labelandchatgpttrainign}
\end{table}

\section{Discussion} 
Our experiments reveal that there are three factors that heavily impact the interpretability of diffusion models. First is the domain shift. Keeping this shift low helps in producing interpretable models. Secondly, keeping the variability high or using strong input signals helps. 
Finally, we revealed that keeping the language encoder frozen heavily impacts the interpretability of diffusion models, which in turn results in worse generative quality in terms of FID. 
This gap is smaller for conditional image generation with Finetune$_F$ being slightly better at generation for six out of eight classes. 
The only two exceptions are ``Edema'' and ``Pneumothorax'', the only two classes of Finetune$_L$ that showed the slightest signs of interpretability. 
This is evidence that Finetune$_L$ did not properly learn to align textual and spatial information for the majority of input prompts. 
Furthermore, we believe this is evidence that focusing on designing interpretable diffusion models could also boost their generative ability and that their interpretability is a reasonable indicator of generative performance. 
Explaining performance differences in medical imaging is of utmost importance as the assessment requires trained experts, whose time is often limited. 
We worked around this by using FID, which has its own limitation due to it being a trained metric and requiring too many samples to be accurate, and conditional generation, which is limited as there is no ground-truth data available. 
Localization of descriptive features on the other hand is more straightforward, as there is a ground truth available. Given that the results shown in our study match with the results of conditional image generation, we believe that it is a valuable metric that should be considered for text-to-image models.
It would be interesting to perform a large-scale study that analyses the quality of images judged by domain experts in relation to FID and CNR to see which metric better captures generative quality.  

\section{Conclusion}
In this paper, we show evidence that the state-of-the-art way of fine-tuning diffusion models to medical tasks results in models that have extraordinary image quality but completely loose interpretability. Although machine learning models with limited interpretability may be suitable for certain industrial applications and entertainment purposes, representation models intended for deployment in medical environments are likely to face significant scrutiny regarding their interpretability in the future.
To alleviate this issue, we perform rigorous experiments that analyze and unveil important relations in terms of training and interpretability of the model.

\section*{Acknowledgements}
This work was supported by HPC resources provided by the Erlangen National High Performance Computing Center (NHR@FAU) of FAU Erlangen-Nürnberg under the NHR projects b143dc and b180dc. NHR funding is provided by federal and Bavarian state authorities. NHR@FAU hardware is partially funded by the German Research Foundation (DFG) – 440719683. Key support was received by the ERC - project MIA-NORMAL 101083647, DFG DC-AIDE 512819079, and by the state of Bavaria. H. Reynaud was supported by Ultromics Ltd. and the UKRI Centre or Doctoral Training in Artificial Intelligence for Healthcare (EP/S023283/1).

\bibliography{aaai24}

\appendix

\newpage

\section{Appendix}
\noindent\textbf{Technical details}
The architecture of our model is based on the $512 \times 512$ pixel model from Stable Diffusion v2.0, as described by \citet{Rombach_2022_CVPR}.
In fine-tuning this model, we automatically fix the latent representation extracted for the localization. As detailed in the Methods section, we extract the attention maps of all cross-attention layers of the U-Net without manually selecting more important layers. 
We note that selectively picking layers might enhance the localization results but we believe that only picking certain layers hurts the overall interpretability. 
The resulting latent dimensions are $S \times D \times T_{max} \times Z \times Z$ where $S$ is defined as the number of reverse diffusion steps and $D$ is defined as the number of cross-attention layers in the encoder and decoder of the U-Net. In our case $D$ is equal to 16. 

\noindent\textbf{Additional experiments on interpretability} First we want to investigate whether the difference in interpretability comes from the domain gap induced by the different impressions. 
MS-CXR impressions are carefully aggregated as opposed to MIMIC impressions which come from a radiology report. 
The result is that parts of the expressions are tokens that were used to anonymize the data and hide private information. Also, they are generally longer and less concise.
Therefore, we repeat our experiment which computes the tokens with the highest localization accuracy for impressions extracted from MIMIC on bounding boxes from MS-CXR. 
The results shown in \cref{tab:mimictokens} indicate that the impressions of the MIMIC dataset are not polished enough to be useful for localization. 
Looking at the results of using MS-CXR impressions in \cref{tab:MSCXRtokens} we can see that for some disease categories, it would be beneficial for localization performance to include other keywords. One example would be the keyword ``enlarged" in the case of Cardiomegaly.  
However, we believe this post hoc analysis would hurt interpretability, which we believe is more important than localization performance.

\label{sec:app_ablation_methods}

\begin{table*}[h!]
\centering

\begin{tabularx}{\linewidth}{Xccccccccccc} 
\toprule
 \multirow{2}{*}{Cat.} & \multicolumn{3}{c}{SDv2}      &&\multicolumn{3}{c}{ Finetune$_{L}$ } && \multicolumn{3}{c}{ Finetune$_{F}$ }\\
 \cmidrule(lr){2-4}\cmidrule(lr){6-8}\cmidrule(lr){10-12}
  & Word  & Occ. & CNR &&  Word  & Occ. & CNR &&  Word  & Occ. & CNR \\
 \midrule 
 \multirow{3}{*}{Atelactasis}& EOS  & 34 & 0.48 &&  EOS & 34& 0.77 && EOS &  34& 0.83 \\ 
 & in  & 28 & -0.03 &&  in & 28& -0.03 && right &  42& 0.78 \\ 
 & right  & 42 & -0.06 &&  right & 42& -0.12 && in &  28& 0.20 \\ 
 \midrule 
 \multirow{3}{*}{Cardiomegaly}& lung  & 71 & 0.68 &&  EOS & 214& 0.32 && moderate &  30& 0.99 \\ 
 & chest  & 36 & 0.44 &&  SOS & 264& 0.15 && bilateral &  44& 0.99 \\ 
 & pulmonary  & 121 & 0.41 &&  vascular & 30& -0.16 && mild &  45& 0.97 \\ 
 \midrule 
\multirow{3}{*}{Consolidation}& pulmonary  & 33 & 0.82 &&  EOS & 58& 0.95 && lung &  39& 1.31 \\ 
 & lung  & 39 & 0.73 &&  a & 29& 0.04 && left &  53& 1.25 \\ 
 & EOS  & 58 & 0.46 &&  to & 50& 0.02 && pulmonary &  33& 1.09 \\ 
 \midrule 
\multirow{3}{*}{Edema}& pulmonary  & 32 & 1.15 &&  EOS & 28& 0.91 && pulmonary &  32& 1.27 \\ 
 & EOS  & 28 & 0.62 &&  of & 42& 0.23 && EOS &  28& 0.80 \\ 
 & of  & 42 & 0.12 &&  the & 64& 0.20 && and &  34& 0.42 \\ 
 \midrule 
\multirow{3}{*}{Lung Opacity}& EOS  & 46 & 0.50 &&  EOS & 46& 0.95 && left &  28& 1.09 \\ 
 & left  & 28 & 0.27 &&  to & 35& 0.06 && right &  45& 0.89 \\ 
 & with  & 33 & 0.15 &&  is & 67& -0.02 && EOS &  46& 0.86 \\ 
 \midrule 
\multirow{3}{*}{Effusion}& SOS  & 91 & 0.10 &&  EOS & 61& 0.41 && left &  56& 0.88 \\ 
 & pulmonary  & 38 & 0.09 &&  SOS & 91& 0.07 && effusion &  26& 0.77 \\ 
 & EOS  & 61 & 0.00 &&  right & 67& -0.12 && pleural &  86& 0.61 \\ 
 \midrule 
 \multirow{3}{*}{Pneumonia}& lung  & 57 & 0.69 &&  EOS & 98& 0.88 && pneumonia. &  52& 1.34 \\ 
 & lobe  & 49 & 0.65 &&  are & 36& -0.01 && lung &  57& 1.33 \\ 
 & pulmonary  & 46 & 0.62 &&  No & 27& -0.09 && left &  72& 1.28 \\ 
 \midrule 
\multirow{3}{*}{Pneumothorax}& Small  & 40 & 0.60 &&  1. & 29& 0.96 && \_\_\_. &  27& 0.75 \\ 
 & catheter  & 29 & 0.60 &&  Small & 40& 0.79 && with &  97& 0.72 \\ 
 & small  & 69 & 0.60 &&  Right & 31& 0.77 && chest &  86& 0.72 \\ 
 \midrule 
\bottomrule
\end{tabularx}
\caption{Number of occurrences and CNR for the three words with the highest localization accuracy of each model on MIMIC. SOS and EOS are start and end of string respectively. Only words with more than 25 occurrences were considered. Radiology reports exceeding the length of the language encoder were ignored.
}
\label{tab:mimictokens}
\end{table*}

\begin{figure}
    \centering
    \includegraphics[width=\linewidth]{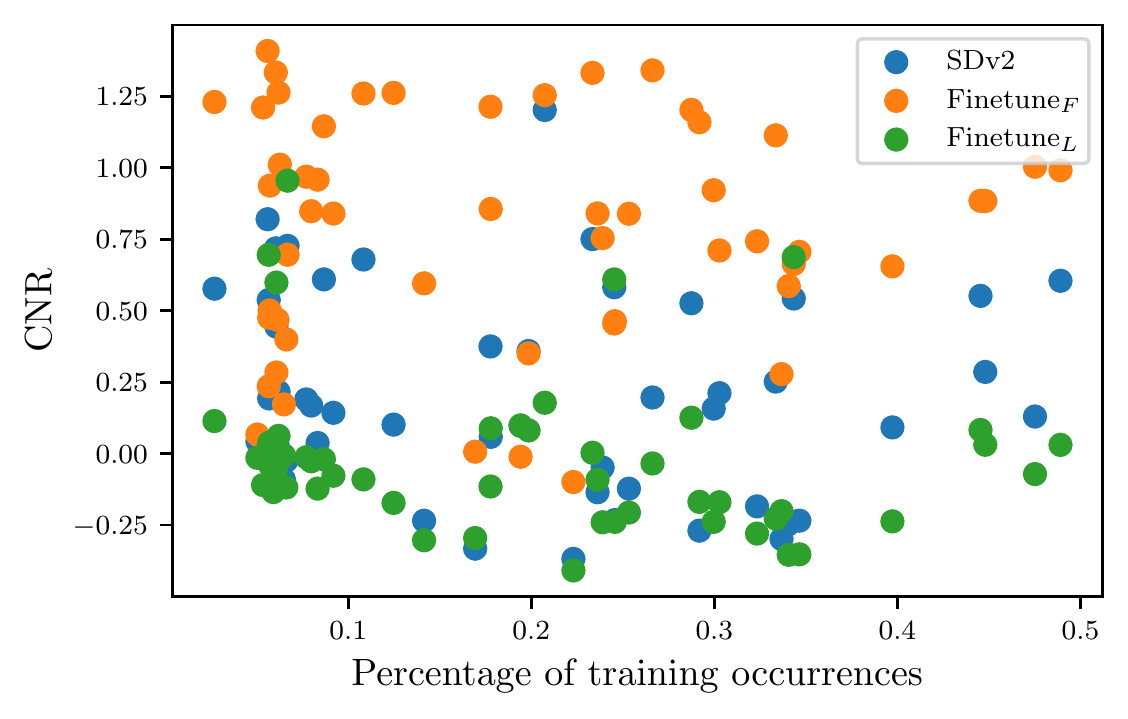}
    \caption{Token experiment: We show the percentage of occurrences of words in impressions of the training data against their ability to localize diseases. We only show words that appear more than 25 times in the validation dataset.}
    \label{fig:TokenExperiment}
\end{figure}

\noindent\textbf{Keywords used for localization} The pipeline described in the method section of the main paper requires that we define the words $y_i$ that we compute the attention over. For MS-COCO we use the object class directly. For the results reported on MS-CXR we define a list of words that we consider as descriptive of the disease class. We summarize the set of words in \cref{tab:attention_keyword mapping}.
This also validates that we could add additional keywords to boost localization performance. Finetune$_F$ only achieved a CNR of 0.67 in our main experiments for Cardiomegaly, but keywords like ``enlarged'' achieve a CNR of 1.16 for this particular disease class. Our initial assumption was that this is based on naturally occurring correlations between certain words in the training dataset with the disease class. Therefore, we analyze whether the number of occurrences of a word in impressions of certain diseases correlates with the localization ability. The results shown in \cref{fig:TokenExperiment} indicate that there is no real correlation visible, which could again be due to the difference between training and validation tokens.

\begin{table*}[]
    \centering
   \begin{tabular}{lrrll}
    \toprule
    \textbf{Category} &$\textbf{N}_D$ & $\textbf{N}_G$ & \textbf{Keyword} & \textbf{Example} \\
    \cmidrule(lr){1-1}\cmidrule(lr){2-2} \cmidrule(lr){3-3}\cmidrule(lr){4-4}\cmidrule(lr){5-5}
    Atelectasis & 61 &  694 & atelectasis, atelectatic & the right lung is atelectatic\\
    Cardiomegaly & 333 &  684& cardiomegaly, cardiac & enlarged cardiac silhouette \\
    Consolidation & 117 &  155& consolidation &right basilar consolidation \\
    Edema &  46 &  408& edema & pulmonary edema \\
    Lung Opacity & 82 & 903& opacity, opacities, opacification & patchy left basilar opacity \\
    Pleural Effusion & 96 & 805 & pleural, effusion, effusions & Small left pleural effusion \\
    Pneumonia &  182 & 271& pneumonia & Right lower lung pneumonia \\
    Pneumothorax & 245 &  182& pneumothorax & trace left pneumothorax \\
    No Finding & 0 &  903 & - & - \\
    \bottomrule
    \end{tabular}
    \caption{Instaces of each disease class in the generative $\textbf{N}_G$ and in the discriminative $\textbf{N}_D$ test dataset. Keywords denote the words that are considered as tokens $\tau_i$ for the attention extraction.}
    \label{tab:attention_keyword mapping}
\end{table*}

\noindent\textbf{Ablation studies} We perform multiple ablation studies on the design choices of our model fine-tuning.
First, we investigate the impact of disabling unconditional training as described in the method section for the frozen (Finetune$_{FnU}$) and the learnable (Finetune$_{LnU}$) approach. We also fine-tune Finetune$_L$ on an additional 10000 steps with frozen language encoder to see if this works as a quick solution to reduce the negative effect of learnable fine-tuning (Finetune$_{60+10}$). Finally, we train Finetune$_{F}$ exclusively on PA images for 30000 steps and a batch size of 128 Finetune$_{PA}$.
Generative results on these ablation studies are summarized in  \cref{tab:ablation_unconditional}. 
Looking at the models without unconditional guidance (Finetune$_{FnU}$ and Finetune$_{FnL}$) we see that the generative results are consistently worse. The FID and FID$_{XRV}$ values drop in all cases indicating the unconditional training is helpful to improve the performance. 
Fine-tuning the learnable model on 10000 additional steps with a frozen encoder has a worse performance than not fine-tuning (Finetune$_{60+10}$ has a FID$_{XRV}$ of 9.3 while Finetune$_F$ has a FID of 7.3).
Finally, we achieve state-of-the-art generative results with Finetune$_{PA}$ which indicates that it might be favourable for the generative model to condition on the view as well as this might improved the peformance.

\begin{table*}
\centering

\begin{tabularx}{\linewidth}{p{2.5cm}ccXccXccXcc} 
\toprule
\multirow{2}{*}{Method}   &\multicolumn{2}{c}{Finetune$_{LnU}$} && \multicolumn{2}{c}{ Finetune$_{FnU}$} && \multicolumn{2}{c}{ Finetune$_{60+10}$} && \multicolumn{2}{c}{Finetune$_{PA}$} \\
 \cmidrule(lr){2-3}\cmidrule(lr){5-6}\cmidrule(lr){8-9}\cmidrule(lr){11-12}
                & Mimic & CXR &&  Mimic & CXR &&  Mimic & CXR &&  Mimic & CXR\\
\cmidrule(lr){2-2}\cmidrule(lr){3-3} \cmidrule(lr){5-5} \cmidrule(lr){6-6}\cmidrule(lr){8-8}\cmidrule(lr){9-9} \cmidrule(lr){11-11}\cmidrule(lr){12-12}

FID  $\downarrow$         & 64.8 & 64.6  &&  88.0  & 85.7  &&  64.8   &  64.7   && 48.0 & 47.9 \\
FID$_{XRV}$ $\downarrow$  & 10.5 & 10.8  &&  16.4  & 16.0  &&  9.5   &  9.3   && 3.5 & 3.6 \\
\bottomrule
\end{tabularx}
\caption{Generative results on our ablation studies of using no unconditional guidance (Finetune$_{FnU}$, Finetune$_{LnU}$), fine-tuning the frozen model (Finetune$_{60+10}$) and training only on PA images (Finetune$_{PA}$). }
\label{tab:ablation_unconditional}
\end{table*}

Next, we evaluate the interpretability of these models by looking at the AUC and Top-1 accuracy in \cref{tab:ablation_localization}. 
The experiments show that unconditional training is beneficial for localization performance. Both training runs with unconditional training enabled (Finetune$_{LnU}$, Finetune$_{FnU}$) are better at localization than the models that did not use it. Finetuning the learnable model on 10000 additional steps with a learnable language encoder did not improve the model noticeably. The interpretability of the Finetune$_{PA}$ model is worse as the test set also contains AP images. 

\begin{table*}[t]
\centering

\begin{tabularx}{\linewidth}{p{2.0cm}ccXccXccXcc} 
\toprule
\multirow{2}{*}{Method}   &\multicolumn{2}{c}{Finetune$_{LnU}$} && \multicolumn{2}{c}{ Finetune$_{FnU}$} && \multicolumn{2}{c}{ Finetune$_{60+10}$} &&  \multicolumn{2}{c}{Finetune$_{PA}$} \\
 \cmidrule(lr){2-3}\cmidrule(lr){5-6}\cmidrule(lr){8-9}\cmidrule(lr){11-12}
  & AUC $\uparrow$  & Top-1 $\uparrow$ && AUC $\uparrow$  & Top-1 $\uparrow$ && AUC $\uparrow$  & Top-1 $\uparrow$ && AUC $\uparrow$ & Top-1 $\uparrow$  \\
\cmidrule(lr){2-2}\cmidrule(lr){3-3} \cmidrule(lr){5-5} \cmidrule(lr){6-6}\cmidrule(lr){8-8}\cmidrule(lr){9-9} \cmidrule(lr){11-11}\cmidrule(lr){12-12}
Atelectasis       & 47.3  & 0.0  && 84.4 &  70.5   && 44.5  & 0.0  && 82.5 & 55.7 \\
Cardiomegaly      & 39.1  & 0.3  && 70.0 &  26.7   && 42.2  & 0.1  && 71.4 & 26.7 \\
Consolidation     & 9.4   & 9.4  && 83.7 &  42.7   && 51.3  & 6.0  && 80.9 & 38.5 \\
Edema             & 55.9  & 41.3 && 87.4 &  67.4   && 54.9  & 30.4 && 86.9 & 86.9 \\
Lung Opacity      & 50.1  & 3.6  && 78.5 &  28.0   && 51.0  & 4.9  && 77.1 & 17.1 \\
Pl. Effusion      & 39.2  & 2.1  && 81.5 &  60.4   && 37.2  & 2.1  && 79.9 & 64.6 \\
Pneumonia         & 47.5  & 1.1  && 84.1 &  61.5   && 47.0  & 2.2  && 81.1 & 35.7 \\
Pneumothorax      & 80.2  & 15.1 && 79.3 &  39.2   && 77.2  & 4.9  && 69.9 & 16.3 \\
 \midrule
Average           & 52.0  & 5.3  && 78.6 &  43.2   && 52.1  & 4.3  && 76.9 & 32.5 \\
\bottomrule
\end{tabularx}
\caption{Ablation study on localization performance.}
\label{tab:ablation_localization}
\end{table*}

\noindent\textbf{Contrast-to-noise ratio ablation study} Finally, we provide some motivation on why we use CNR, instead of the more commonly used CNR$_{abs}$. We compute CNR using the following formula: 
\begin{equation}
 CNR= \frac{(\mu_A - \mu_{\overline{A}})}{
(\sigma_{A}^2 + \sigma^2_{\overline{A}}) ^ {\frac{1}{2}}
} 
\end{equation}
where $A$ describes the ground-truth segmentation or bounding box and $\overline{A}$ its complement.
However, in literature it is more common to use 
\begin{equation}
CNR_{abs} = \frac{|\mu_A - \mu_{\overline{A}}|}{
(\sigma_{A}^2 + \sigma^2_{\overline{A}}) ^ {\frac{1}{2}}
}
\end{equation}
However, this means that if the localization is terrible, and the mean value of the attention outside the region of interest is higher than inside, then the numerator would be positive and the localization would count towards positive results. 
In \cref{tab:cnrabs} we compare both approaches for our three main models. It is clear that there is no meaningful difference between both approaches when the localization results are generally good (Funitune$_F$). However, for SDv2 and Finetune$_L$ CNR$_{abs}$ overestimates the performance visibly. 

\begin{table*}
\centering
\begin{tabularx}{0.8\textwidth}{rcclccccccccc}
\toprule
  &  & Method  & \multicolumn{1}{l}{Atel.} & \multicolumn{1}{l}{Card.} & \multicolumn{1}{l}{Cons.} & \multicolumn{1}{l}{L-Op.} & \multicolumn{1}{l}{Edem.} & \multicolumn{1}{l}{Pnem.} & \multicolumn{1}{l}{Pnth.} & \multicolumn{1}{l}{P-Ef.} & \multicolumn{1}{l}{Avg.} \\ 
\cmidrule(lr){4-4} \cmidrule(lr){5-5}\cmidrule(lr){6-6}\cmidrule(lr){7-7}\cmidrule(lr){8-8}\cmidrule(lr){9-9}\cmidrule(lr){10-10}\cmidrule(lr){11-11}\cmidrule(lr){12-12} 
 \multirow{3}{*}{\rotatebox{90}{CNR}} &  & SDv2 & 0.23   & 0.14   & 0.21   & 0.37   & 0.87   & 0.34   & 0.64   & 0.01   & 0.321 \\ 
& & Finetune$_{L}$ & -0.20  & -0.20  & -0.01  & 0.00   & 0.11  & -0.08  & 0.75   & -0.35  & 0.075 \\ 
& & Finetune$_{F}$& 1.30   & 0.74   & 1.29   & 1.07   &1.34   & 1.20   & 0.59   & 1.09   & 0.942 \\ 
\midrule
 \multirow{3}{*}{\rotatebox{90}{CNR$_{abs}$}} & & SDv2 & 0.37 & 0.37 & 0.31 & 0.41 & 0.87 & 0.41 & 0.65 & 0.23  & 0.445 \\ 
& & Finetune$_{L}$ & 0.30 & 0.27 & 0.26 & 0.30 & 0.20 & 0.25 & 0.78 & 0.39 & 0.386 \\ 
& & Finetune$_{F}$ & 1.30   & 0.74   & 1.29   & 1.07   & 1.34   & 1.20   & 0.59   & 1.09   & 0.944 \\ 
\bottomrule
\end{tabularx}
\caption{
Localization results using $\text{CNR}_{abs}$ and CNR on a single run.}
\label{tab:cnrabs}
\end{table*}

\noindent \textbf{ChatGPT impressions} To get impressions from ChatGPT we first prompt the model to ``Give me three radiography reports of a chest x-ray of a lung with [disease class]". 
The results are close to the structure in MIMIC, which leads to the assumption that some reports of MIMIC were part of the training dataset. 
Next, we ask the model only to return the impression section of the reports, as we did as a preprocessing step for the training on the real reports.  
Finally, we ask the model to generate 100 of those with emphasis on variability because we observed that if we do not ask for that, the model will simply return the same 3-5 impressions. 
The following shows five examples:

\begin{enumerate}
    \item Right middle lobe atelectasis with associated volume loss and mediastinal shift. Clinical correlation and further evaluation recommended.
    \item Left lower lobe atelectasis with associated volume loss. Clinical correlation and further evaluation recommended to determine the underlying cause.
    \item Complete atelectasis of the right upper lobe with associated volume loss and mediastinal shift. Clinical correlation and further evaluation recommended to determine the underlying cause and appropriate management.
    \item Atelectasis involving the left upper lobe with associated crowding of adjacent lung parenchyma. Clinical correlation advised.
    \item Right lower lobe atelectasis with associated volume loss. Further evaluation recommended for proper management.
\end{enumerate}

\end{document}